\documentclass[unnumsec,webpdf,contemporary,large]{oup-authoring-template}%

\usepackage{graphicx}
\usepackage{booktabs}
\graphicspath{{Fig/}}
\usepackage{tabularx}
\usepackage{xurl}
\usepackage{subfig}
\usepackage[table]{xcolor}
\usepackage{tcolorbox}
\usepackage{xcolor}
\usepackage[normalem]{ulem}

\theoremstyle{thmstyleone}%
\theoremstyle{thmstyletwo}%
\theoremstyle{thmstylethree}%

\begin{document}

\journaltitle{Briefings in Bioinformatics}
\DOI{DOI added during production}
\copyrightyear{2026}
\pubyear{2026}
\vol{XX}
\issue{x}
\access{Published: Date added during production}
\appnotes{Review}

\firstpage{1}


\title[Short Article Title]{Do Papers Tell the Whole Story? A Benchmark and Framework for Uncovering Hidden Implementation Gaps in Bioinformatics}

\author[1,2]{Tianxiang Xu}
\author[1,2]{Xiaoyan Zhu}
\author[1,2]{Xin Lai}
\author[1,2]{Xin Lian}
\author[1,2]{Sizhe Dang}
\author[1,2]{Hangyu Cheng}
\author[1,2,$\ast$]{Jiayin Wang\ORCID{0000-0002-3862-6557}}

\address[1]{\orgdiv{School of Computer Science and Technology}, \orgname{Xi'an Jiaotong University}, \orgaddress{\postcode{710049}, \state{Xi'an}, \country{China}}}
\address[2]{\orgdiv{Shaanxi Engineering Research Center of Medical and Health Big Data}, \orgname{Xi'an Jiaotong University}, \orgaddress{\postcode{710049}, \state{Xi'an}, \country{China}}}

\corresp[$\ast$]{Corresponding author. \href{wangjiayin@mail.xjtu.edu.cn}{wangjiayin@mail.xjtu.edu.cn}}

\received{Date}{0}{Year}
\revised{Date}{0}{Year}
\accepted{Date}{0}{Year}



\abstract{As bioinformatics software is increasingly applied across a broader range of scenarios and the rapid development of large language models (LLMs) further lowers the barriers to software use and development, the composition of the bioinformatics research community is undergoing substantial change. Consequently, a growing number of researchers require a deeper understanding of methodological details and software behavior. In this context, systematically analyzing the relationship between paper descriptions and code implementations is emerging as an important new challenge in the field. To address this challenge, we introduce paper-code consistency analysis as a new research perspective and construct BioCon, the first benchmark dataset for paper-code consistency analysis in bioinformatics. Furthermore, we develop a unified cross-modal analysis framework to systematically investigate this problem from three perspectives: sentence-level detection, cross-modal retrieval, and project-level assessment. Experimental results demonstrate that the proposed framework can effectively model the semantic relationships between scientific publications and software implementations. Further case studies reveal that paper-code inconsistency is not a single phenomenon but arises from multiple underlying causes, among which Author-Perceived Non-Essential Details represents the most prevalent category. These findings suggest that paper-code consistency analysis is not merely a technical problem but also raises broader discussions regarding knowledge dissemination, the boundaries of code disclosure, and community norms. We hope that this work will encourage the bioinformatics community to re-examine the relationship between scientific publications and software implementations while providing a foundation for future research in paper-code consistency analysis.}


\boxedtext{Key Messages}{
\begin{itemize}
\item We introduce paper-code consistency analysis as a new research perspective and discuss its significance in the era of LLMs.
\item We construct BioCon, the first benchmark for paper-code consistency analysis in bioinformatics, and establish a unified analytical framework to facilitate future research.
\item We systematically evaluate the potential of LLMs for paper-code consistency analysis from sentence-level, retrieval-level, and project-level perspectives.
\item  Through case studies and error analysis, we reveal the potential causes and practical manifestations of paper-code inconsistency in real-world bioinformatics software.
\end{itemize}}

\maketitle


\section{Introduction}
Bioinformatics, as a critical interdisciplinary field integrating modern biology and information technology, has become an indispensable foundation for life sciences research \cite{ma2024bioinformatics, attwood2019global}. With the completion of the Human Genome Project and the rapid advancement of high-throughput sequencing technologies, the continuous accumulation of massive biological data has significantly accelerated progress in bioinformatics research \cite{liu2024bioinformatics}. Against this backdrop, an increasing number of bioinformatics research outputs are disseminated not only through scientific publications but also through open-source software repositories \cite{sharma2024analytical}, and are widely applied in practical scenarios such as genomic analysis, genetic disease screening, cancer diagnosis and treatment, and precision medicine \cite{wolde2025current,abbas2025advancing,thornton2021alphafold}. Software is no longer merely a by-product of scientific research; rather, it has increasingly become an important medium for knowledge dissemination and application in bioinformatics \cite{martin2024fairsoft}.

Meanwhile, the user population of bioinformatics software is undergoing significant transformation. Traditionally, such software primarily served professional researchers with strong computational backgrounds and domain expertise, who were generally capable of understanding software implementation logic by reading papers, examining source code, and leveraging their domain knowledge. However, as bioinformatics expands beyond specialized research settings into broader application domains, its user base has gradually extended to clinicians, healthcare practitioners, and interdisciplinary researchers. In particular, with the rapid advancement of artificial intelligence (AI) technologies and LLMs, an increasing number of researchers without formal training in bioinformatics are becoming involved in the development and use of bioinformatics analysis workflows. While the barriers to software usage are decreasing, the demand for understanding methodological details, parameter configurations, implementation logic, and algorithmic behavior continues to grow among these users \cite{Lavazza2023empirical}.

In the field of bioinformatics, scientific publications and software code play distinct yet complementary roles. Publications typically focus on algorithmic design, theoretical foundations, and experimental results, whereas code translates these methods into executable software systems and contains numerous implementation-level details, including parameter settings, preprocessing procedures, and engineering optimization strategies \cite{ziemann2023five,Thimbleby2024Improving}. For the growing number of researchers without extensive bioinformatics backgrounds, relying solely on publications often fails to provide complete implementation information, while analyzing code alone may not adequately convey the methodological motivations and theoretical foundations behind the software. Consequently, establishing correspondence between paper descriptions and code implementations has become increasingly important for understanding and using bioinformatics software.

In fact, this demand is already widespread within the bioinformatics community. For well-known software projects such as AlphaFold \cite{jumper2021highly}, Seurat \cite{hao2024dictionary} and STAR \cite{dobin2013star}, users frequently raise questions in community forums and GitHub Issues\footnote{https://github.com/google-deepmind/alphafold/issues/1071}\footnote{https://github.com/satijalab/seurat/issues/10297}\footnote{https://github.com/alexdobin/STAR/issues/2684} regarding issues such as which code components implement a specific step described in a paper, where default parameters are defined, or whether a particular algorithmic detail is executed exactly as described in the publication. These questions do not necessarily imply the existence of software defects or errors between papers and code. Rather, they reflect the growing need among an expanding user population to understand the correspondence between scientific publications and software implementations. Implementation details that may be implicitly understood by domain experts are often far less obvious to users from diverse backgrounds. Therefore, systematically analyzing the relationship between papers and code, and helping researchers understand the actual implementation process of software, is emerging as an important new challenge.

We argue that, driven by these ongoing changes in bioinformatics, the relationship between papers and code is extending beyond the traditional concept of reproducibility \cite{baykal2024genomic} toward broader issues involving software understanding, knowledge dissemination, and scientific traceability. As software is increasingly reused, extended, and deployed in diverse application settings, researchers are becoming concerned not only with whether software can be executed successfully, but also with how methods are implemented, which details are explicitly documented, which knowledge remains embedded in the code, and how different forms of knowledge are transferred between papers and software implementations.

Motivated by this observation, we propose paper-code consistency analysis as a new research perspective for systematically analyzing the semantic correspondence between methodological descriptions in scientific publications and software implementations. It is important to emphasize that paper-code consistency analysis is not intended to determine whether software is correct, nor is it designed to identify scientific misconduct. Instead, it aims to help researchers better understand software implementations and the underlying processes of knowledge representation by analyzing inconsistencies between papers and code and exploring their potential causes.

Nevertheless, research on paper-code relationships remains at an early stage. On the one hand, the emergence of LLMs has made automated consistency analysis increasingly feasible. On the other hand, the bioinformatics community still lacks systematic studies regarding the definition, boundaries, and evaluation criteria of paper-code consistency. Therefore, before deeper discussions can be conducted, it is necessary to establish a unified data foundation and analytical framework.

To this end, we construct BioCon, the first benchmark dataset for paper-code consistency analysis in bioinformatics, and propose a unified cross-modal framework for modeling semantic consistency between paper descriptions and software implementations. Based on this benchmark, we systematically investigate paper-code consistency analysis from three perspectives: sentence-level detection, cross-modal retrieval, and project-level assessment. Furthermore, we conduct case studies on real-world bioinformatics software projects to investigate the underlying causes and manifestations of paper-code inconsistency in practical scenarios.

Specifically, this study seeks to answer the following research questions:
\begin{enumerate}[1.]
\item At the sentence level, can the framework accurately determine the consistency between paper descriptions and code implementations?
\item From a retrieval perspective, can the framework effectively locate corresponding implementation units from large-scale code repositories given a paper description?
\item At the project level, can the framework provide an overall consistency assessment based on sentence-level predictions?
\item What are the potential causes and manifestations of paper-code inconsistency in real-world bioinformatics software?
\end{enumerate}

Based on these research questions, this study not only demonstrates the feasibility of automated paper-code consistency analysis but also reveals the underlying causes of inconsistencies between paper descriptions and code implementations in real-world bioinformatics software. For example, in some projects, developers may consider certain implementation details as non-essential information and therefore choose not to fully disclose the corresponding code. Such inconsistencies may reflect different understandings between software developers, researchers, and users regarding knowledge dissemination and software development practices, for which a unified consensus has not yet been established. As bioinformatics applications continue to expand and user demands continue to evolve, important questions remain regarding which information should be considered essential implementation knowledge, which aspects should be consistently represented across papers and code, and how appropriate community standards should be established. We hope that this work will not only advance automated consistency analysis techniques but also encourage the bioinformatics community to reconsider the boundaries of implementation disclosure, mechanisms of knowledge dissemination, and the development of consistency standards.

\section{Related Work}
In recent years, AI and large-scale pretrained models have achieved remarkable progress, demonstrating strong performance in various tasks, including natural language processing, code retrieval, and code generation \cite{zhang2024code, liu2025empirical, huang2024code}. These advances have made it increasingly feasible to automatically analyze the relationship between scientific publications and software implementations. However, to the best of our knowledge, existing studies have not systematically investigated the consistency between paper descriptions and code implementations. Instead, related research has primarily focused on three adjacent areas: research reproducibility, code-text alignment, and software quality assessment.

\textbf{Research Reproducibility.} Research on reproducibility primarily investigates whether scientific findings can be independently reproduced, including aspects such as the availability of datasets and source code, the reconstructability of experimental environments, and the executability of analytical workflows \cite{kapoor2023Leakage, Trisovic2022large}. For example, Siddiq et al. \cite{siddiq2025large} analyzed papers published in the software engineering domain over the past five years and found that only 13.3\% of papers were free from reproducibility-related issues.

However, such studies typically consider whether experiments can be reproduced as the primary objective, focusing mainly on software while paying relatively limited attention to the corresponding scientific publications. Importantly, successful reproduction does not necessarily indicate consistency between methodological descriptions and code implementations. A paper may omit critical implementation steps, simplify methodological details, or even contain discrepancies with the actual implementation. These issues may not directly result in reproduction failure, but they can affect how subsequent researchers understand the underlying methodology, correctly utilize the software, and further extend or reuse the approach. Therefore, the objective of paper-code consistency analysis differs fundamentally from traditional reproducibility research. Rather than determining whether an experiment can be repeated, it focuses on analyzing the semantic correspondence between knowledge representations in scientific publications and their actual realization in software implementations.

\textbf{Code-Text Alignment.} In software engineering, extensive research has been conducted on modeling the relationship between natural language descriptions and source code \cite{zhang2026survey}, including tasks such as code retrieval \cite{li2025pseudobridge, Husain2019CodeSearchNetCE} and code generation \cite{zhao2026towards}. These studies generally assume that a meaningful correspondence exists between natural language descriptions and code, aiming to learn cross-modal representations to improve retrieval accuracy or code generation performance \cite{zan-etal-2023-large}. However, the problem investigated in this work is fundamentally different. Instead of asking which code segment is most relevant to a given textual description, paper-code consistency analysis further asks: Is the scientific sentence described in the paper actually implemented in the corresponding code?

Even when existing retrieval models successfully identify relevant code fragments, they cannot determine whether the corresponding paper description contains omissions, deviations, or inaccuracies. In other words, code retrieval and related tasks aim to discover potential associations between text and code, whereas paper-code consistency analysis aims to verify the authenticity, completeness, and semantic alignment of such relationships. Therefore, paper-code consistency analysis is not a direct extension of existing code retrieval tasks but represents a fundamentally different research problem. Consequently, existing software engineering methods and datasets cannot be directly transferred to this emerging task without additional considerations.

\textbf{Software Quality Assessment.} In the bioinformatics domain, previous studies have extensively investigated software quality from various perspectives, including usability, maintainability, documentation completeness, coding practices, and long-term sustainability \cite{martin2024fairsoft, zhu2025reaching, ferenc2022empirical, Coelho2024For}. These studies have significantly advanced software engineering practices for bioinformatics applications. However, existing software quality assessments primarily focus on the engineering quality of software itself, while paying limited attention to the relationship between software implementations and scientific publications as a means of knowledge dissemination. As bioinformatics software continues to expand from specialized research environments to broader application scenarios, including clinical practice and interdisciplinary collaborations, an increasing number of users need to rely on both publications and source code to understand software behaviors. Under these circumstances, whether papers and code remain consistent is no longer merely a software engineering concern, but also involves broader issues related to the accuracy of scientific knowledge representation and reliable knowledge dissemination.

\section{Benchmark Construction}
\begin{figure*}[htbp]
  \centering
  \includegraphics[width=1\linewidth]{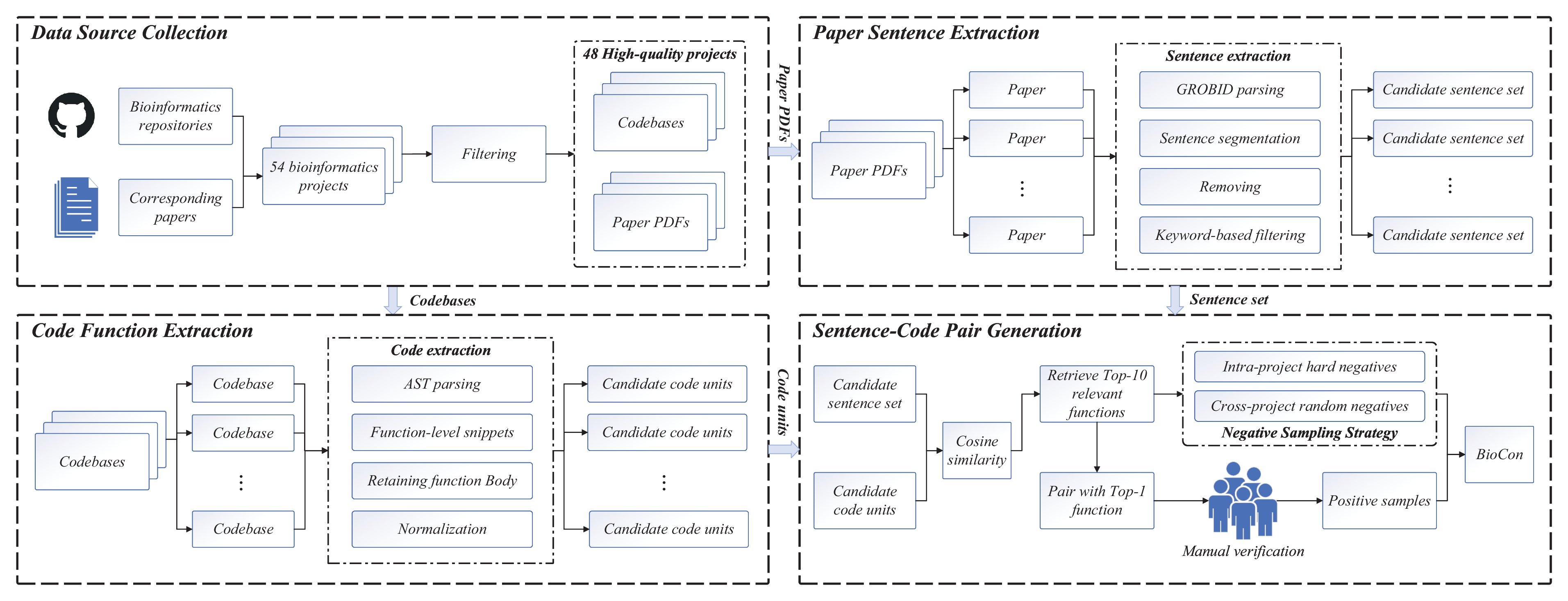}
  \caption{Overview of the BioCon benchmark construction pipeline.}
  \label{fig:1}
\end{figure*}

To systematically study the problem of paper-code consistency analysis, we construct a benchmark dataset for the bioinformatics domain, termed BioCon. This section presents the construction pipeline in detail, including data source collection, paper sentence extraction, code function extraction, and sentence-code pair generation. The overall workflow is illustrated in Figure~\ref{fig:1}.

\subsection{Data Source Collection}
We begin by collecting bioinformatics software projects along with their corresponding research papers. Following the selection criteria established in prior work \cite{zhu2025reaching}, we identify 54 bioinformatics projects from GitHub that are implemented in Python and are explicitly associated with published papers. The full-text versions of the corresponding publications are then retrieved.

To ensure data quality, we perform rigorous manual filtering. First, all papers are carefully reviewed, and three projects lacking clearly defined methodological sections are excluded. Next, all code repositories are analyzed, and three additional projects with parsing failures or significant structural issues are removed. As a result, 48 high-quality bioinformatics software projects are retained, each containing both a complete paper (PDF) and a functional codebase.

Statistical details are summarized in Table~\ref{tab:1}, including average code size, number of files, number of functions, cyclomatic complexity, as well as project popularity (measured by stars) and paper citation counts. The statistics indicate that these projects exhibit substantial code scale and structural complexity, rendering manual consistency analysis impractical in real-world scenarios. Furthermore, in terms of impact (stars and citations), the selected projects are representative and influential within the domain, providing a reliable foundation for the paper-code consistency analysis task. More detailed information about each project is provided in Supplementary Section A.

\begin{table}[htbp]
  \caption{Summary statistics of the 48 bioinformatics software projects in BioCon. All values represent the average statistics calculated across the 48 projects.}
  \centering
  \begin{tabular*}{\columnwidth}{@{\extracolsep\fill}cccc@{\extracolsep\fill}}
    \toprule
    \textbf{Metrics}    & \textbf{Value} & \textbf{Metrics} & \textbf{Value} \\
    \midrule
        Lines of Code    &  19134    &  Cyclomatic Complexity  & 2732\\
        Number of Files   & 76    &  Number of Stars  & 355   \\
        Number of Functions    &  681    &  Citation Count  & 603     \\
    \botrule
  \end{tabular*}
  \label{tab:1}
\end{table}

\subsection{Paper Sentence Extraction}
To obtain text units relevant to code implementation, we extract sentence-level content from the full text of each paper. Specifically, we first employ GROBID to perform structured parsing of PDF documents, enabling accurate extraction of textual content along with section information. GROBID is chosen due to its optimization for scholarly documents, allowing reliable identification of sections such as \textit{Methods} and \textit{Experiments}.

The extracted text is then segmented into sentence-level units. Based on section metadata, we remove content unrelated to implementation, including references, appendices, conclusions, and acknowledgments. However, even within retained sections, some sentences may still be irrelevant (e.g., background descriptions or result analysis). Therefore, we further apply keyword-based filtering to prioritize sentences closely related to methodological implementation, thereby improving the effectiveness of subsequent sentence-code matching.

Ultimately, for each paper, we obtain a set of high-quality candidate sentences, which are treated as potential methodological descriptions and will be aligned with code functions in subsequent steps.

\subsection{Code Function Extraction}
Given the software repositories, we perform function-level parsing of source code to construct semantically meaningful code units. Specifically, Python source code is analyzed using abstract syntax trees (AST), from which function-level code snippets are extracted.

Function-level granularity strikes a balance between semantic completeness and manageable input length, making it well-suited for the paper-code consistency analysis task. For each function, we retain the full function body as the primary representation, along with auxiliary metadata such as function name, file path, and comments. These additional features provide useful semantic cues that facilitate both model training and human annotation.

We also apply standard normalization procedures, such as removing empty lines and redundant formatting symbols, to ensure consistency of input sequences. As a result, each software project is represented as a collection of functions, where each function serves as a potential implementation unit for subsequent sentence-code alignment.

\subsection{Sentence-Code Pair Generation}
After obtaining the sets of paper sentences and code functions, we construct sentence-code pairs as samples for the consistency analysis task. For each sentence, we first retrieve the top-10 most semantically similar functions from the corresponding project using cosine similarity. This step reduces the candidate space and lowers the cost of manual annotation.

Next, each sentence is paired with its top-1 candidate function and evaluated by three domain experts (from bioinformatics, software engineering, and computer science). A pair is labeled as consistent only if all three experts unanimously agree that the function correctly implements the functionality described in the sentence. All other cases are excluded from the labeled dataset to minimize noise introduced by subjective judgment. Detailed annotation procedures and representative examples are provided in Supplementary Section A.

This process yields a set of high-confidence positive samples, ensuring clear semantic correspondence between paper descriptions and code implementations while avoiding noise from irrelevant combinations.

\subsection{Negative Sampling Strategy}
To construct negative samples and enhance the discriminative capability of the framework, we design a hybrid negative sampling strategy that combines hard negatives and random negatives. For each positive sentence-code pair, negative samples are generated from two sources:
\begin{itemize}
\item \textbf{Intra-project hard negatives}: Functions from the same project that are semantically related but inconsistent with the sentence (e.g., functions ranked from Top-5 to Top-10 in cosine similarity retrieval). These samples exhibit high semantic similarity but do not implement the described functionality, thereby improving the model’s ability to distinguish fine-grained semantic differences.
\item \textbf{Cross-project random negatives}: Functions randomly sampled from other projects and paired with the sentence. This strategy simulates cross-project semantic noise and increases the diversity and realism of the data distribution.
\end{itemize} 

For each sentence, we construct two hard negatives and three random negatives. Compared to random negatives, hard negatives are more challenging and prevent the model from relying on superficial lexical cues, thereby improving its ability to capture deeper semantic consistency. This ratio provides sufficiently challenging supervision while maintaining adequate sample diversity during training. We further conduct ablation experiments by varying the proportion of hard negatives and random negatives. The results demonstrate that the 2:3 configuration achieves stable and competitive performance across multiple evaluation levels, and is therefore adopted as the default negative sampling strategy. Detailed experimental results are provided in Supplementary Section B.

\subsection{Benchmark Statistics}
Following the above pipeline, we construct the BioCon benchmark dataset, consisting of paper sentences, function-level code snippets, and corresponding consistency labels. Each sample in BioCon contains three components: a paper sentence, a code function, and a binary consistency label.

To ensure a realistic evaluation setting, the dataset is split at the project level to prevent data leakage. Specifically, software projects in the test set are entirely unseen during training, enabling robust assessment of cross-project generalization. BioCon is divided into training, validation, and test sets with a ratio of 8:1:1. Detailed statistics are reported in Table~\ref{tab:2}.

\begin{table}[htbp]
  \caption{Dataset statistics of BioCon. The projects included in different splits are mutually exclusive, ensuring that no data leakage occurs.}
  \centering
  \begin{tabular*}{\columnwidth}{@{\extracolsep\fill}ccccc@{\extracolsep\fill}}
    \toprule
    \textbf{Split}    & \textbf{\# Projects} & \textbf{\# Consistent} & \textbf{\# Inconsistent} & \textbf{\# Total} \\
    \midrule
        Train	& 38&	957&	4785&	5742\\
        Validation&	5&	90&	450&	540\\
        Test&	5&	83&	415&	498  \\
    \midrule
        Total&	48&	1130&	5650&	6780     \\
    \botrule
  \end{tabular*}
  \label{tab:2}
\end{table}

\section{Method}
\subsection{Problem Formulation}
paper-code consistency analysis aims to characterize the semantic alignment between methodological descriptions in papers and their corresponding software implementations. Formally, given a sentence $s$ from a paper and a function-level code snippet $c$, the goal is to determine whether the code $c$ correctly implements the functionality described by $s$.

To this end, we define a unified cross-modal consistency function:
\begin{equation}
    f(s, c) \rightarrow p
\end{equation}
where $p\in[0,1]$ denotes the probability that the sentence-code pair is consistent. This function serves not only as a binary classifier at the sentence level but also as a scoring function for ranking and aggregation, thereby supporting multi-level consistency analysis.

Based on this formulation, the core objective of this work is to learn a function $f$ that effectively models semantic consistency between natural language and code, enabling downstream multi-dimensional evaluation tasks.

\subsection{Cross-modal Consistency Framework}
\begin{figure*}[t]
  \centering
  \includegraphics[width=1\linewidth]{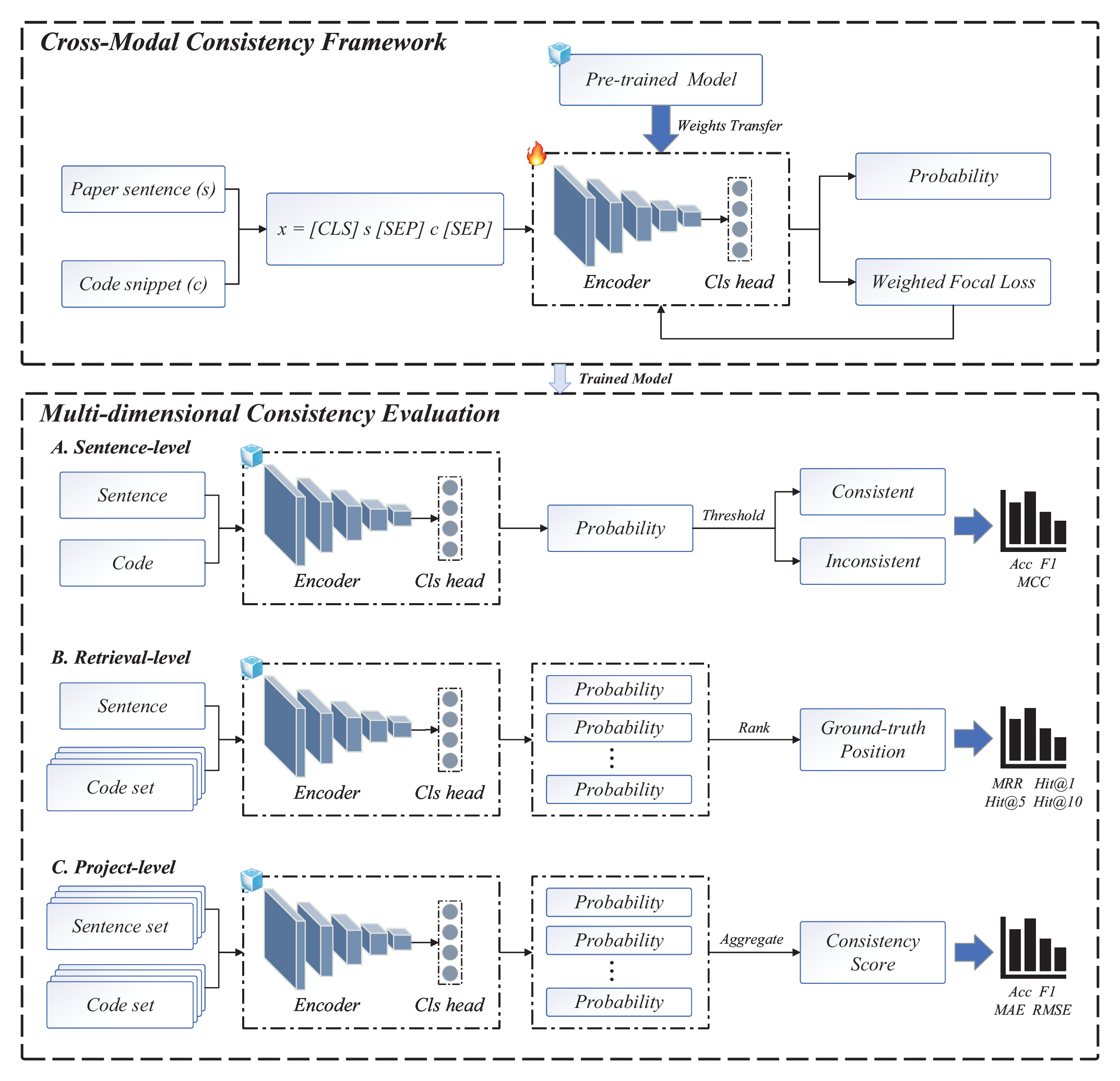}
  \caption{Overall architecture of the proposed cross-modal consistency analysis framework.}
  \label{fig:2}
\end{figure*}

To effectively model consistency between bioinformatics papers and code, we propose a cross-modal consistency analysis framework based on pre-trained models, as illustrated in Figure \ref{fig:2}.

\textbf{Input Representation.} For each sentence-code pair $(s,c)$, we construct a unified input sequence:
\begin{equation}
    x=[CLS]\,sentence\, [SEP]\,code\,[SEP]
\end{equation}
where $[CLS]$ is the classification token and $[SEP]$ separates the sentence and code. This formulation enables the model to jointly encode natural language and code within a shared context, facilitating cross-modal interaction.

\textbf{Encoder.} We adopt UniXcoder \cite{guo2022unixcoder} as the backbone encoder. UniXcoder is a pre-trained model designed for code representation learning, trained on large-scale corpora of code and natural language, and is capable of capturing semantic alignment between the two modalities. Our framework is model-agnostic and can be easily extended to other pre-trained models such as CodeT5+ \cite{wang2023codet5+} and CodeBERT \cite{feng2020codebert}.

Given the input sequence $x$, the encoder produces contextualized hidden representations:
\begin{equation}
    H=Encoder(x),\;\;H\in\mathbb{R}^{T\times d}
\end{equation}
where $T$ is the sequence length and $d$ is the hidden dimension.

To obtain a global semantic representation of the sentence-code pair, we use the hidden vector corresponding to the first token ($[CLS]$):
\begin{equation}
    h=H_{CLS}
\end{equation}
which captures the overall semantic relationship between the paper sentence and the code function.

\textbf{Classification Head.} Based on the global representation h, a linear layer is applied to produce the consistency probability:
\begin{equation}
    p=\operatorname{Softmax}(Wh+b)
\end{equation}
where $W$ and $b$ are learnable parameters, and $p \in [0,1]$ represents the predicted consistency probability.

\subsection{Training Objective}
Due to class imbalance in BioCon, directly applying standard cross-entropy loss may bias the model toward the majority class, thereby weakening its ability to identify minority-class samples. To address this issue, we adopt a weighted Focal Loss to emphasize hard and minority samples.

The standard cross-entropy loss is defined as:
\begin{equation}
    \mathcal{L}_{CE}=-\log(p_y)
\end{equation}
where $p_y$ denotes the predicted probability for the ground-truth label $y$.

Focal Loss \cite{lin2017focal} introduces a modulating factor to reduce the contribution of easily classified samples:
\begin{equation}
    \mathcal{L}_{FL}=-(1-p_y)^{\gamma}\log(p_y)
\end{equation}
where $\gamma$ is the focusing parameter controlling the emphasis on hard samples.

To further address class imbalance, a weighting factor $\alpha$ is incorporated, leading to the final weighted Focal Loss:
\begin{equation}
    \mathcal{L}=-\alpha(1-p_y)^{\gamma}\log(p_y)
\end{equation}

By minimizing this loss, the model learns more discriminative representations under imbalanced conditions, improving its ability to detect semantic consistency in real-world bioinformatics scenarios.

\subsection{Multi-dimensional Consistency Evaluation}
Building upon the unified modeling framework, we further evaluate paper-code consistency from multiple perspectives to comprehensively assess model capability.

\textbf{Sentence-level Evaluation.} At the sentence level, consistency analysis is formulated as a binary classification task. For a given pair $(s,c)$, the model outputs a probability $p$, which is converted into a discrete label $y \in \{0,1\}$ using a threshold $T$. This evaluation measures the model’s ability to capture local semantic consistency.

\textbf{Retrieval-level Evaluation.} At the retrieval level, given a sentence $s$ and a candidate code set $C=\{c_1,c_2,...,c_n\}$, the model computes consistency scores for all candidates and ranks them accordingly. The position of the ground-truth implementation in the ranked list is then used for evaluation. This setting assesses the model’s ability to locate relevant implementation units within large-scale code repositories.

\textbf{Project-level Evaluation.} At the project level, we aggregate sentence-level predictions to characterize the overall consistency of a bioinformatics project. For a project $P$ with a set of sentence-code pairs, the overall consistency score is defined as:
\begin{equation}
    Score(P)=Aggregate(f(s_1,c_1),f(s_2,c_2),...,f(s_n,c_n))
\end{equation}
where $Aggregate$ denotes an aggregation function (e.g., mean, ratio, or maximum). This score provides a global view of consistency between the paper and its corresponding codebase.

\section{Experiments}
\subsection{Experimental Setup}
\subsubsection{Encoders}
To evaluate the effectiveness of the proposed framework and the BioCon benchmark, we compare several representative pre-trained models as backbone encoders, including CodeBERT \cite{feng2020codebert}, UniXcoder \cite{guo2022unixcoder}, CodeT5+ \cite{wang2023codet5+}, Qwen3 \cite{yang2025qwen3} and CodeGen \cite{nijkamp2022codegen}. These models have demonstrated strong performance in code understanding and cross-modal learning tasks, making them suitable baselines for this study.

To ensure a fair comparison, all models adopt a unified input format, where the paper sentence and code function are concatenated into a single sequence. Considering differences in maximum input length across encoders, the sequence length is uniformly limited to 512 tokens. For sequences exceeding this limit, the code portion is truncated first to preserve the complete sentence, thereby avoiding the loss of critical semantic information. In addition, all models are trained using the weighted Focal Loss, with the focusing parameter $\gamma=2$ and class weights $\alpha=[1.0,5.0]$, to mitigate the impact of class imbalance in the dataset.

\subsubsection{Implementation details}
All experiments are conducted on eight NVIDIA RTX 4090 GPUs. Pre-trained model weights are loaded via Hugging Face \cite{wolf2020transformers}. We use the AdamW optimizer with a learning rate of $2\times10^{-5}$, a batch size of 16, and train each model for 10 epochs. Early stopping is applied, and the best model is selected based on validation performance. To reduce randomness, all reported results are averaged over three runs with different random seeds. Due to differences in model sizes, the training time per epoch ranges approximately from 1 min 36 s to 3 min 14 s.

\subsubsection{Evaluation Metrics}
To comprehensively assess the framework, we adopt multiple evaluation metrics tailored to different task settings.

\textbf{Sentence-level classification.} In addition to Accuracy (Acc) and macro-F1 (F1), we report the Matthews Correlation Coefficient (MCC) \cite{chicco2021matthews,guo2017calibration} as a key metric. Unlike Acc and F1, MCC provides a balanced evaluation by considering all elements of the confusion matrix, making it particularly robust under class imbalance. Acc and F1 range from $[0,1]$, while MCC ranges from $[-1,1]$.

\textbf{Cross-modal retrieval.} To evaluate the ability to locate correct implementations within candidate sets, we use Mean Reciprocal Rank (MRR) and Hit@k ($k \in \{1,5,10\}$). MRR measures the relative ranking position of the correct code, assigning higher scores to higher ranks, with values in $[0,1]$. Hit@k evaluates whether the correct implementation appears within the top-$k$ retrieved results.

\textbf{Project-level consistency.} For project-level evaluation, we consider both classification and regression perspectives. The aggregated consistency score is thresholded at 0.5 to produce discrete labels, evaluated using Acc and F1. Additionally, accounting for the inherent uncertainty of ground-truth consistency boundaries, we report Mean Absolute Error (MAE) and Root Mean Square Error (RMSE) to quantify the deviation between predicted and ground-truth inconsistency ratios. Both metrics range from $[0, +\infty)$, with lower values indicating better performance.

\subsubsection{Evaluation Protocols}
To ensure consistency and reproducibility, all evaluations are conducted on the BioCon test set under unified settings.
\begin{enumerate}[1.]
\item \textbf{Sentence-level classification.} Each sample is a sentence-code pair $(s,c)$, evaluated directly without candidate construction.
\item \textbf{Intra-project retrieval.} For each test sentence, the candidate set includes all functions within the same project.
\item \textbf{Cross-project retrieval.} The candidate set is expanded to include all functions across all test projects.
\item \textbf{Project-level consistency.} Based on five test projects, we construct three variants with different inconsistency distributions: 10$\%$ inconsistent samples (mixed), 30$\%$ inconsistent samples (mixed), and fully consistent (clean). This results in 20 project instances for evaluation.
\end{enumerate}

\subsection{Sentence-level Results}
Table~\ref{tab:3}  presents the sentence-level classification results of different encoders on the BioCon test set. All models are trained under identical settings, differing only in the choice of encoder.

\begin{table}[!t]
\setlength\tabcolsep{5pt}
\renewcommand{\arraystretch}{1.4}
  \caption{Sentence-level consistency classification performance of different pre-trained models. All values and corresponding errors are reported as the mean and standard deviation over three runs with different random seeds. The best and second-best results are highlighted in \textbf{bold} and \underline{underlined}, respectively.}
  \centering
  \begin{tabular*}{\columnwidth}{@{\extracolsep\fill}cccc@{\extracolsep\fill}}
    \toprule
    \textbf{Encoder}    & \textbf{Acc ($\uparrow$)} & \textbf{F1 ($\uparrow$)} & \textbf{MCC ($\uparrow$)} \\
    \midrule
        CodeBERT&0.8668$_{\pm 0.0041}$ &0.7350$_{\pm 0.0117}$	&\underline{0.4852}$_{\pm0.0102}$\\
        CodeT5+&	0.8541$_{\pm 0.0018}$	&\underline{0.7380}$_{\pm 0.0054}$ 	&0.4763$_{\pm 0.0109}$ 	\\
        CodeGen&	0.8534$_{\pm 0.0065}$ &	0.7260$_{\pm 0.0108}$ &0.4531$_{\pm 0.0220}$ 	\\
        Qwen3& \underline{0.8728}$_{\pm 0.0048}$ &	0.7303$_{\pm 0.0017}$ &0.4829$_{\pm 0.0108}$ \\
        UniXcoder&	\textbf{0.8949}$_{\pm 0.0064}$&	\textbf{0.7736}$_{\pm 0.0139}$&	\textbf{0.5754}$_{\pm 0.0272}$\\
    \botrule
  \end{tabular*}
  \label{tab:3}
\end{table}

Overall, all pre-trained models achieve strong performance, indicating that paper-code consistency analysis can be effectively modeled via cross-modal representation learning. This also validates the feasibility of the BioCon dataset and task formulation.

Among all models, UniXcoder consistently achieves the best performance across all metrics. Compared to CodeBERT, it improves Acc by 0.0281 and F1 by 0.0386, with a notable gain of 0.0902 in MCC, highlighting its superior discriminative ability under imbalanced conditions. This advantage can be attributed to UniXcoder’s unified representation learning mechanism, which more effectively captures deep semantic alignment between natural language and code. Interestingly, while other models exhibit relatively small gaps in Acc and F1 compared to UniXcoder, the differences in MCC are more pronounced. This suggests that although these models perform well in overall classification accuracy, they are less effective in distinguishing minority-class samples under imbalanced settings.

\subsection{Retrieval Results}
To evaluate the framework in cross-modal retrieval scenarios, we conduct experiments under both intra-project and cross-project settings, with results reported in Table~\ref{tab:4} and~\ref{tab:5}.

\begin{table}[!t]
\setlength\tabcolsep{2pt}
\renewcommand{\arraystretch}{1.4}
  \caption{Comparison of retrieval performance under the intra-project setting. All values and corresponding errors are reported as the mean and standard deviation over three runs with different random seeds. The best and second-best results are highlighted in \textbf{bold} and \underline{underlined}, respectively.}
  \centering
  \begin{tabular*}{\columnwidth}{@{\extracolsep\fill}ccccc@{\extracolsep\fill}}
    \toprule
    \textbf{Encoder}    & \textbf{MRR ($\uparrow$)} & \textbf{Hit@1 ($\uparrow$)} & \textbf{Hit@5 ($\uparrow$)} & \textbf{Hit@10 ($\uparrow$)}\\
    \midrule
        CodeBERT&0.5205$_{\pm 0.0072}$ &0.4217$_{\pm 0.0120}$	&\underline{0.6546}$_{\pm0.0106}$ &0.6908$_{\pm0.0080}$\\
        CodeT5+&	0.5059$_{\pm 0.0044}$	&0.4096$_{\pm 0.0070}$ 	&0.6346$_{\pm 0.0040}$ 	&\underline{0.6988}$_{\pm 0.0070}$\\
        CodeGen&	\underline{0.5362}$_{\pm 0.0057}$ &	\underline{0.4458}$_{\pm 0.0139}$ &0.6305$_{\pm 0.0106}$ 	&0.6707$_{\pm 0.0080}$\\
        Qwen3& 0.5346$_{\pm 0.0098}$ &	\textbf{0.4618}$_{\pm 0.0145}$ &0.6104$_{\pm 0.0080}$ &0.6627$_{\pm 0.0120}$\\
        UniXcoder&	\textbf{0.5452}$_{\pm 0.0040}$&	0.4297$_{\pm 0.0106}$&	\textbf{0.7108}$_{\pm 0.0070}$ &\textbf{0.7389}$_{\pm 0.0040}$\\
    \botrule
  \end{tabular*}
  \label{tab:4}
\end{table}

\begin{table}[!t]
\setlength\tabcolsep{2pt}
\renewcommand{\arraystretch}{1.4}
  \caption{Comparison of retrieval performance under the cross-project setting. All values and corresponding errors are reported as the mean and standard deviation over three runs with different random seeds. The best and second-best results are highlighted in \textbf{bold} and \underline{underlined}, respectively.}
  \centering
  \begin{tabular*}{\columnwidth}{@{\extracolsep\fill}ccccc@{\extracolsep\fill}}
    \toprule
    \textbf{Encoder}    & \textbf{MRR ($\uparrow$)} & \textbf{Hit@1 ($\uparrow$)} & \textbf{Hit@5 ($\uparrow$)} & \textbf{Hit@10 ($\uparrow$)}\\
    \midrule
        CodeBERT&\textbf{0.3558}$_{\pm 0.0069}$&	\underline{0.2851}$_{\pm 0.0080}$&	\textbf{0.4257}$_{\pm 0.0145}$ &\textbf{0.5341}$_{\pm 0.0040}$\\
        CodeT5+&	0.3180$_{\pm 0.0053}$	&0.2570$_{\pm 0.0040}$ 	&0.4016$_{\pm 0.0080}$ 	&0.4699$_{\pm 0.0070}$\\
        CodeGen&	0.2968$_{\pm 0.0053}$ &	0.2450$_{\pm 0.0106}$ &0.3695$_{\pm 0.0106}$ 	&0.4659$_{\pm 0.0040}$\\
        Qwen3& 0.3353$_{\pm 0.0194}$ &	\textbf{0.2892}$_{\pm 0.0184}$ &0.3655$_{\pm 0.0080}$ &0.4377$_{\pm 0.0145}$\\
        UniXcoder&	\underline{0.3509}$_{\pm 0.0041}$ &	0.2771$_{\pm 0.0070}$ &\underline{0.4217}$_{\pm 0.0070}$ 	&\underline{0.5261}$_{\pm 0.0040}$\\
    \botrule
  \end{tabular*}
  \label{tab:5}
\end{table}

Under the intra-project setting, all models achieve relatively strong performance, indicating that they can capture sentence-code alignment when the candidate space is restricted. Among all models, UniXcoder achieves the best results in terms of MRR (0.5452), Hit@5 (0.7108), and Hit@10 (0.7389), demonstrating its ability to rank the correct implementation near the top in most cases.

In contrast, Qwen3 achieves the highest Hit@1 (0.4618) but performs slightly worse on Hit@5 and Hit@10, suggesting a tendency toward more aggressive ranking behavior. Specifically, the model assigns high confidence to a limited number of candidates but exhibits relatively weaker overall ranking stability. Overall, UniXcoder and CodeBERT exhibit more balanced performance across metrics, reflecting stronger robustness in ranking.

Under the cross-project setting, performance drops significantly for all models. For example, UniXcoder’s MRR decreases from 0.5452 to 0.3509 (-35.6$\%$), while CodeBERT drops from 0.5205 to 0.3558 (-31.6$\%$). Similar declines are observed for Hit@1, Hit@5, and Hit@10. This performance degradation is not solely caused by the increased number of candidate functions but also results from the absence of project-specific semantic and structural priors in cross-project retrieval. In intra-project retrieval, consistent naming conventions, coding styles, domain-specific terminology, and function organization patterns provide additional contextual information, thereby reducing semantic ambiguity among candidate functions. In contrast, cross-project retrieval introduces substantial variations in implementation strategies, abstraction levels, and function organization patterns. As a result, similar functionalities may be implemented through different function structures across projects, while semantically similar functions may serve different purposes in different software systems, leading to increased semantic ambiguity.

Interestingly, although UniXcoder achieves the best performance under the intra-project setting, CodeBERT slightly outperforms UniXcoder under the cross-project setting in terms of MRR (0.3558) and Hit@10 (0.5341). This observation suggests that retrieval performance depends not only on local semantic matching capability but also on model robustness to implementation variations across different projects. The intra-project and cross-project settings represent two distinct challenges in paper-code consistency analysis. Intra-project retrieval primarily evaluates whether models can perform semantic alignment within a relatively consistent project context. In contrast, cross-project retrieval requires models to generalize across substantial differences in coding conventions, function decomposition strategies, abstraction levels, and project-specific contexts. Therefore, cross-project retrieval is not merely an expanded candidate search problem; rather, it introduces additional challenges in cross-project semantic generalization and better reflects practical deployment scenarios.

\begin{figure*}[t]
\centering
\includegraphics[width=1\linewidth]{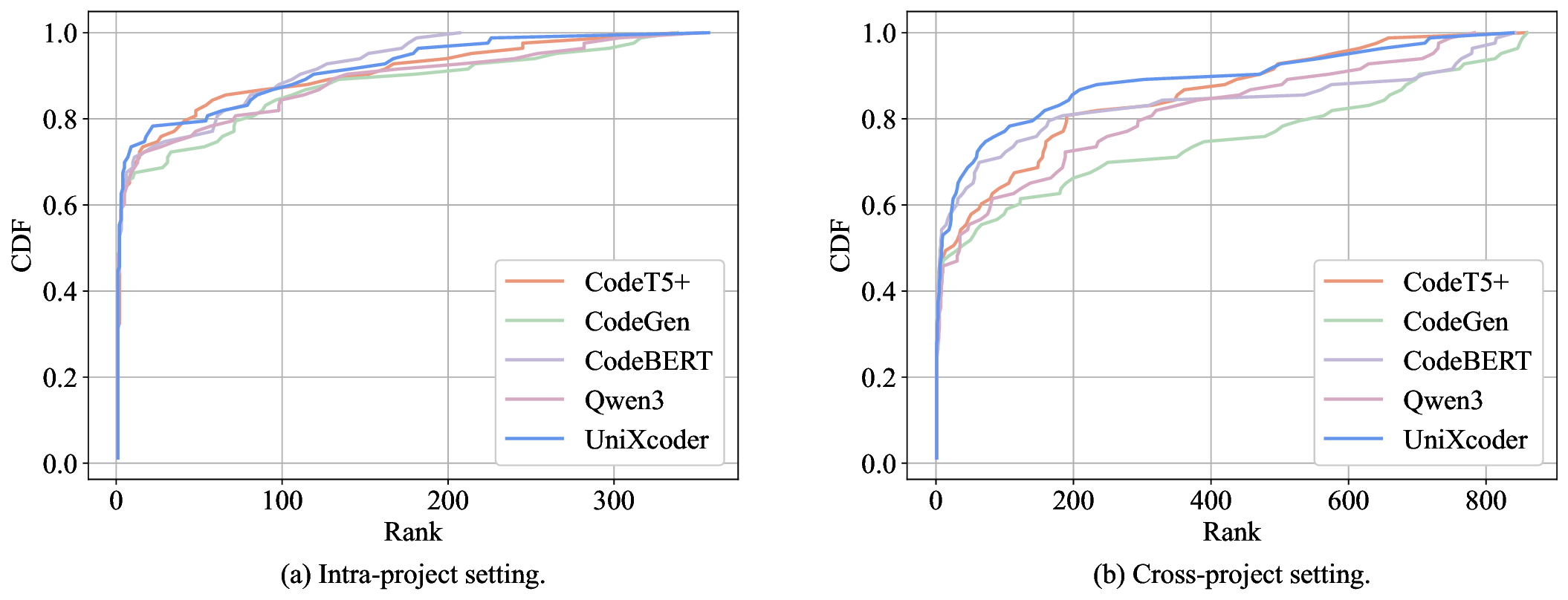}
\caption{Cumulative distribution function curves of ranking results. Panels (a) and (b) present the CDF curves of five models under the intra-project and cross-project settings, respectively. In each CDF curve, the x-axis represents the ranking position, while the y-axis indicates the cumulative proportion of queries whose retrieved implementation units have ranks less than or equal to the corresponding value.}
\label{fig:3}
\end{figure*}

To further analyze retrieval performance, we visualize the ranking distributions using cumulative distribution function (CDF) curves, as shown in Figure~\ref{fig:3}.

Under the intra-project setting, the CDF curves are steep, indicating that most correct matches are ranked within a small range. UniXcoder consistently dominates within the top-30 ranks, further confirming its superiority in ranking quality. In contrast, under the cross-project setting, all curves shift to the right, indicating a significant decline in ranking quality. Moreover, the performance gaps between models become more pronounced, suggesting that some models struggle under more challenging retrieval conditions.

Overall, retrieval evaluation provides a stricter assessment than sentence-level classification. While existing models perform well in constrained candidate spaces, they exhibit clear limitations in cross-project scenarios. Future work should focus on improving cross-project generalization, for example, by incorporating structured code representations or leveraging larger-scale cross-domain pre-training data, to better support real-world bioinformatics applications.

\subsection{Project-level Results}
To assess paper-code consistency from a holistic perspective, we further conduct project-level consistency analysis. Specifically, based on sentence-level predictions, we aggregate all sentence-code pairs within the same project to obtain an overall consistency score for each software project.

Table~\ref{tab:6} reports the project-level performance of different encoders using mean aggregation. Overall, all models demonstrate relatively stable performance in the project-level consistency assessment task, with UniXcoder achieving the best results across three evaluation metrics. Specifically, UniXcoder obtains an Acc of 0.8167 while achieving the lowest errors in terms of MAE (0.2792) and RMSE (0.3759). These results indicate that, beyond strong sentence-level discrimination, UniXcoder can effectively capture the overall consistency of a project.

\begin{table}[!t]
\setlength\tabcolsep{2pt}
\renewcommand{\arraystretch}{1.4}
  \caption{Project-level consistency evaluation across different models. All values and corresponding errors are reported as the mean and standard deviation over three runs with different random seeds. The best and second-best results are highlighted in \textbf{bold} and \underline{underlined}, respectively.}
  \centering
  \begin{tabular*}{\columnwidth}{@{\extracolsep\fill}ccccc@{\extracolsep\fill}}
    \toprule
    \textbf{Encoder}    & \textbf{Acc ($\uparrow$)} & \textbf{F1 ($\uparrow$)} & \textbf{MAE ($\downarrow$)} & \textbf{RMSE ($\downarrow$)}\\
    \midrule
        CodeBERT&0.7833$_{\pm 0.0167}$	&0.6191$_{\pm 0.0112}$ 	&\underline{0.2825}$_{\pm 0.0219}$ 	&\underline{0.3771}$_{\pm 0.0279}$\\
        CodeT5+&	0.7667$_{\pm 0.0167}$	&0.6055$_{\pm 0.0215}$ 	&0.3023$_{\pm 0.0077}$ 	&0.3942$_{\pm 0.0170}$\\
        CodeGen&	0.7833$_{\pm 0.0167}$ &	0.5943$_{\pm 0.0136}$ &0.3130$_{\pm 0.0146}$ 	&0.4027$_{\pm 0.0228}$\\
        Qwen3& \underline{0.8000}$_{\pm 0.0289}$ &	\textbf{0.6632}$_{\pm 0.0397}$ &0.2869$_{\pm 0.0139}$ &0.3815$_{\pm 0.0172}$\\
        UniXcoder&	\textbf{0.8167}$_{\pm 0.0167}$&	\underline{0.6520}$_{\pm 0.0441}$&	\textbf{0.2792}$_{\pm 0.0054}$ &\textbf{0.3759}$_{\pm 0.0073}$\\
    \botrule
  \end{tabular*}
  \label{tab:6}
\end{table}

By jointly analyzing sentence-level and project-level results, an important observation emerges: although predictions for individual sentence-code pairs may contain errors, appropriate aggregation can still yield accurate project-level consistency estimation. This suggests that paper-code consistency analysis exhibits a noise-tolerant property, where local prediction errors can be mitigated at the global level. In other words, occasional misclassifications at the sentence level are effectively smoothed during aggregation, ensuring stable overall assessment. This finding has practical implications. In real-world bioinformatics scenarios, researchers are often more concerned with the overall reliability of a project rather than the correctness of individual sentence-code alignments. Therefore, project-level consistency provides a more application-oriented evaluation perspective.

\section{Case Study}
\subsection{Overall Results}

To investigate the causes and implications of paper-code inconsistency in real-world bioinformatics software, we conduct a case study on publicly available bioinformatics projects. Specifically, we select articles from the most recent issues of three representative journals, namely \textit{Bioinformatics}, \textit{Genome Biology} and \textit{PLOS Computational Biology}, as analysis targets. Among these studies, the code links provided in five papers were no longer accessible, while the remaining 44 projects released codebases implemented in Python. Therefore, the following analyses are conducted based on these 44 real-world projects.

For each project, we automatically extract sentences related to methods or implementation following the pipeline described earlier, and pair them with all functions in the corresponding code repository. The trained framework is then used to compute consistency scores. For each sentence, we retrieve the top-10 most relevant functions as candidates and apply the following decision rules:
\begin{enumerate}[1.]
\item If none of the top-10 candidates has a high-confidence match (i.e., all consistency scores are below a threshold of 0.5), the sentence is considered potentially inconsistent with the code;
\item Otherwise, the sentence is considered potentially consistent.
\end{enumerate}

It should be noted that this process does not rely on manually constructed sentence-function correspondences. Instead, it simulates the process by which researchers locate implementation details from paper descriptions when examining software in practice. Therefore, the focus of this section is not whether the software is implemented correctly, but whether a clear correspondence can be established between paper descriptions and publicly available code. 

\begin{figure*}[htbp]
  \centering
  \includegraphics[width=0.98\linewidth]{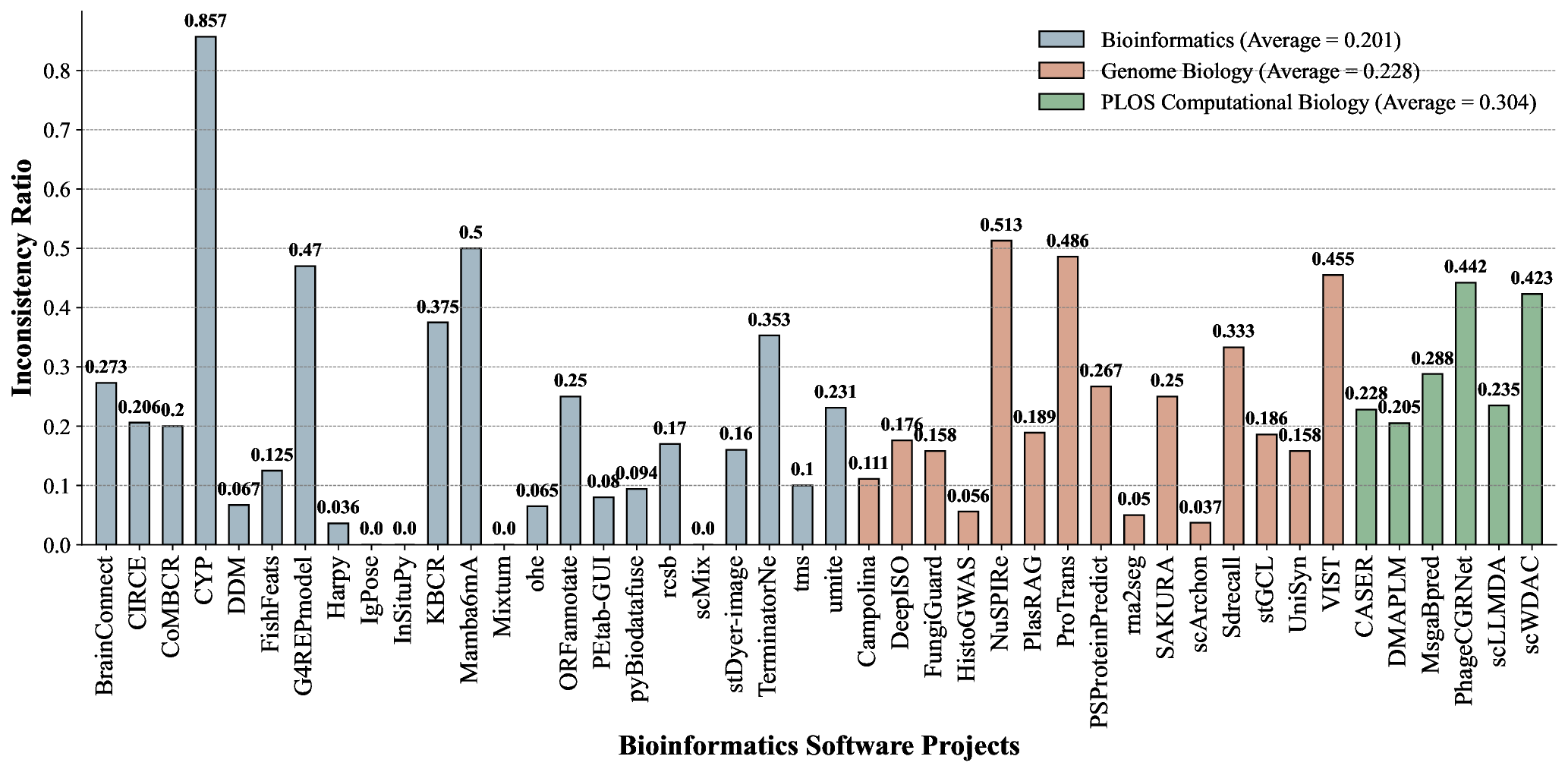}
  \caption{Inconsistency ratios across 44 real-world bioinformatics software projects. The selected projects were collected from the latest issues of Bioinformatics, Genome Biology, and PLOS Computational Biology, including 23, 15, and 6 projects, respectively. All values are rounded to three decimal places.}
  \label{fig:4}
\end{figure*}

Based on the above analysis, we further calculated the paper-code consistency ratio, inconsistency ratio, and average consistency score for each project. The detailed project statistics are provided in Supplementary Table S7 (Supplementary Section C), while Figure~\ref{fig:4} summarizes the inconsistency ratios across the 44 projects. We observed substantial variation in paper-code relationship characteristics across projects. Some projects \cite{rombaut2026scalable,wirth2026insitupy} exhibited relatively high consistency levels, indicating that most methodological descriptions could be clearly mapped to corresponding implementations in the code. In contrast, other projects \cite{kim2026prediction,rosignoli2026deep} showed relatively high inconsistency ratios, suggesting potential discrepancies between paper descriptions and code implementations. Further analysis revealed that projects published in \textit{Bioinformatics} and \textit{Genome Biology} exhibited lower average inconsistency ratios than those published in \textit{PLOS Computational Biology}. 

Overall, the average observable inconsistency ratio among the 44 projects is approximately 22.4\%, suggesting that most methodological descriptions can be mapped to corresponding implementations in publicly available repositories, while localized inconsistencies remain prevalent. However, this estimation should be interpreted with caution. First, the current analysis only includes projects whose code remains publicly accessible, whereas five projects with unavailable code links were excluded from automated analysis. From the perspective of paper-code consistency, inaccessible code itself represents a fundamental barrier to verifying whether methodological descriptions can be confirmed at the implementation level. Therefore, such projects may suffer from more severe limitations in terms of implementation traceability and verifiability. Second, the analyzed projects were collected from three highly recognized bioinformatics journals, and these publications may generally represent projects with relatively stronger research practices, software maintenance, and code availability standards. Consequently, the current sample may not fully represent the broader bioinformatics software ecosystem. Taken together, the observed inconsistency ratio of 22.4\% should be regarded as a conservative estimate among currently accessible projects published in high-quality bioinformatics journals. The actual prevalence of paper-code inconsistency in the broader bioinformatics software landscape may be higher.

Furthermore, similar discrepancies between scientific descriptions and software implementations have also been reported in other research domains, including clinical research \cite{cro2020evidence} and software engineering \cite{siddiq2025large}. Although these studies adopt different definitions and evaluation methodologies from ours, their findings suggest that the relationship between scientific descriptions and computational implementations may represent a broader challenge across research disciplines rather than an issue unique to bioinformatics.

\subsection{Human Validation}
Since the analysis of 44 projects primarily relies on the automated consistency framework, we further conduct an independent human validation study to evaluate the reliability of the obtained results. Specifically, we select 10 projects from the three journals for validation, including four projects from \textit{Bioinformatics}, four from \textit{Genome Biology}, and two from \textit{PLOS Computational Biology}. For each selected project, we manually review the framework predictions and compare them with human assessment results. It should be noted that these projects are not included in the BioCon dataset and are therefore not involved in model training or testing. Consequently, this independent validation provides an additional evaluation of the framework’s generalization capability on unseen real-world projects.

As shown in Table~\ref{tab:7}, across the 10 independently validated projects, the framework achieves an average Acc of 0.7858, with an F1 and MCC of 0.7552 and 0.5757, respectively. These results indicate that the proposed framework can reliably identify potential consistency relationships between paper descriptions and code implementations, although its predictions are not completely error-free. In particular, false-positive cases indicate that some samples classified as inconsistent by the framework may actually have corresponding implementations consistent with the descriptions in the papers. Therefore, the estimated inconsistency ratio derived from large-scale analysis should be interpreted as an approximation of the potential inconsistency level rather than an exact measurement without uncertainty.

\begin{table}[!t]
\setlength\tabcolsep{4pt}
\renewcommand{\arraystretch}{1.4}
  \caption{Comparison between framework predictions and human validation results for 10 bioinformatics software projects. For each project, evaluation metrics are calculated based on the predictions and manual verification results of all candidate sentences within that project. The last row reports the average performance across the 10 projects.}
  \centering
  \begin{tabular*}{\columnwidth}{@{\extracolsep\fill}cccc@{\extracolsep\fill}}
    \toprule
    \textbf{Project}    & \textbf{Acc ($\uparrow$)} & \textbf{F1 ($\uparrow$)} & \textbf{MCC ($\uparrow$)} \\
    \midrule
        Project 1  & 0.9286 & 0.9048 & 0.8257 \\
Project 2  & 0.7778 & 0.7750 & 0.5500 \\
Project 3  & 0.5000 & 0.5000 & 0.1636 \\
Project 4  & 0.7692 & 0.7068 & 0.5394 \\
Project 5  & 0.7500 & 0.7334 & 0.5774 \\
Project 6  & 0.9333 & 0.9283 & 0.8660 \\
Project 7  & 0.8500 & 0.8466 & 0.6975 \\
Project 8  & 0.7368 & 0.7077 & 0.5443 \\
Project 9  & 0.6923 & 0.6389 & 0.3175 \\
Project 10 & 0.9200 & 0.8106 & 0.6757 \\
\midrule
\textbf{Average} & \textbf{0.7858$_{\pm 0.0420}$} & \textbf{0.7552$_{\pm 0.0405}$} & \textbf{0.5757$_{\pm 0.0676}$} \\
    \botrule
  \end{tabular*}
  \label{tab:7}
\end{table}

Combined with the overall analysis of the 44 projects, this independent validation further supports the feasibility of applying the framework to large-scale investigations of paper-code consistency in real-world bioinformatics software. It should be acknowledged that the current human validation is limited in sample size, and different projects vary substantially in terms of code complexity, biological domains, and implementation strategies. Therefore, future studies should perform broader human evaluations on larger and more diverse project collections to further assess the framework’s generalization capability across different bioinformatics tasks and complex software environments.

\subsection{Potential Causes of paper-code Inconsistency and Framework Error Analysis}
\subsubsection{Potential Causes of paper-code Inconsistency}

We further analyze the inconsistent cases identified during human validation to understand the underlying causes of discrepancies between paper descriptions and code implementations in real-world bioinformatics software. It should be noted that the analysis is based on specific inconsistent cases identified from the 10 manually validated projects described above. We summarize these cases into three potential causes of paper-code inconsistency. For each category, we present three representative cases, as summarized in Table~\ref{tab:8}.

\textbf{Operational Issues During Development and Release.} This category accounts for 26.67\% of the observed inconsistencies. These cases mainly arise from operational issues during software development, version management, or public release processes. Specifically, certain implementations described in the paper may exist during the development stage but are absent from the final publicly released repository due to issues such as incomplete synchronization, incorrect version uploading, or insufficient code maintenance. For example, one project \cite{kim2026prediction} described in the paper that model training was performed using a composite loss function consisting of BCE Loss and Weighted Positive Loss. However, the implementation of Weighted Positive Loss was not found in the publicly available repository. After communication with the authors, they confirmed that an earlier version of the code had been mistakenly uploaded, resulting in only the BCE Loss implementation being released while the Weighted Positive Loss component was missing. Such cases reflect potential limitations in current bioinformatics software development and release workflows, particularly regarding version management, code organization, and long-term maintenance. Since software development often involves multiple rounds of experimentation and iterative modification, discrepancies may emerge between the final released code and the specific implementation version used for reported experiments, resulting in paper-code inconsistency.

\textbf{Author-Perceived Non-Essential Details.} This category represents the largest proportion of observed inconsistencies (55\%) and primarily reflects differences in authors’ perceptions regarding which components should be publicly disclosed. Specifically, certain elements, including training procedures, dataset preprocessing pipelines, auxiliary analyses, or parameter configurations, may be considered by authors as non-essential components of the software and therefore are not included in the released code. For example, some projects \cite{rosignoli2026deep, tjaden2026terminatornet} provide detailed descriptions of training strategies and data preprocessing procedures in their publications; however, their publicly available repositories only contain trained models and processed datasets, without the corresponding training or preprocessing code. Importantly, such cases do not necessarily indicate that the authors failed to implement these components. Instead, they may reflect different interpretations of the minimum publicly available implementation required for software disclosure. For some researchers, training scripts, preprocessing workflows, auxiliary analyses, and parameter configurations may be regarded as experimental details rather than essential software components that must be released. This phenomenon highlights a broader question regarding modern bioinformatics software: which types of implementation details should be considered essential knowledge and therefore require public disclosure? Currently, no unified consensus exists regarding this issue. As the user community of bioinformatics software continues to expand beyond traditional domain experts, establishing clearer community expectations for implementation disclosure remains an important topic for further discussion.

\textbf{Limited Code Accessibility Due to Resource Constraints.} This category accounts for 18.33\% of the observed inconsistencies and is mainly associated with constraints arising from research environments and software ecosystems. Specifically, some implement-ations depend on resources that cannot be directly released, including internal servers, private computational environments, or external software components. Consequently, these components are not included in publicly available repositories. For example, one project \cite{rosignoli2026deep} described a filtering strategy in the paper, but the corresponding implementation was not identified in the released code. After communication with the authors, we learned that this component was executed within a server environment and was not included in the public repository. Similarly, some components in another project \cite{hua2026self} relied on external libraries or existing frameworks and therefore did not contain explicit implementations within the project repository. These cases indicate that paper-code inconsistency does not always originate from errors or omissions. Instead, it may result from modular software dependencies and resource limitations within real-world software ecosystems. Nevertheless, for researchers attempting to independently reproduce or extend these methods, such hidden dependencies may still create practical barriers.

\begin{table*}[t]
  \setlength\tabcolsep{7pt}
  \renewcommand{\arraystretch}{1.5}
  \caption{Representative cases of paper-code inconsistencies identified by the framework in real-world bioinformatics software and their potential causes. Since a single project may involve multiple causes of inconsistency, the reported proportions represent the percentage of inconsistent sentences rather than the percentage of projects. Note that the three categories are not mutually exclusive; each inconsistent sentence is assigned to the most likely cause based on manual analysis.}
  \centering
  \begin{tabularx}{\textwidth}{@{} >{\hsize=0.5\hsize}X >{\hsize=1.7\hsize}X c >{\hsize=0.8\hsize}X   @{}}
    \toprule
    \multicolumn{1}{c}{\textbf{Cause}} & \multicolumn{1}{c}{\textbf{Paper Sentence}} & \textbf{Probability} & \multicolumn{1}{c}{\textbf{Human Validation}} \\
    \midrule
    Operational Issues During Development and Release &	During the generation of pseudo labels, a Spies Capture Rate (SCR) assesses the model's performance of each bin in identifying the known positives of "spies" from unlabeled dataset.&0.1345 &Inconsistent. No implementation related to SCR was identified in the code. \\
    (26.67\%)& The weighted positive loss term penalizes the errors of the predictions in truly positive instances, while the BCE term trains the model using entire samples as a binary classifier.&0.1733&Inconsistent. The implementation of weighted positive loss was not identified in the code. \\
    &The whole slide images were segmented into smaller patches of 256 × 256 pixels using a sliding window approach.&0.1749&Inconsistent. The image size used in the code is 112 × 112.\\
    \midrule
    Author-Perceived Non-Essential Details (55\%)&Model parameters were determined by Adam stochastic optimization (Kingma 2014) for 1000 iterations with a batch size of 200 using the binary cross-entropy loss function.&0.1401& Inconsistent. No implementation of Adam optimizer and BCE loss was identified in the code.\\
    &The second step, homology reduction, a critical process in preparing input data for a neural network model (Li et al.2023) , aimed to further eliminate redundancy, thereby minimizing bias, reducing overfitting risk, and enhancing the ability of the model to generalize to unseen data.&0.2033&Inconsistent. No implementation of homology reduction was identified in the code.\\
    &Thus, we interrogated the relative contributions of different features toward the model's output using SHAP (SHapley Additive exPlanations) analysis (Sundararajan and Najmi 2020).&0.4503&Inconsistent. No usage or implementation related to SHAP was identified in the code.\\
    \midrule
    Limited Code Accessibility Due to Resource Constraints&Only windows where G $\geq$ T pass through for further RG4-binding analysis, producing a focused set of candidate regions defined by both structural flexibility and enriched G-S-Y-F-R content.&0.1072&Inconsistent. The filtering strategy based on G $\geq$ T for sliding windows was not implemented in the code.\\
    (18.33\%)&The encoder's output, along with the positional information of the masked patches, is passed to the decoder to reconstruct the masked content.&0.1696&Inconsistent. No implementation of the corresponding functionality was identified in the code.\\
    &The multi-head self-attention mechanism enables the model to weigh the relevance of different input elements in the input set when learning global representations.&0.4345&Inconsistent. No customized attention implementation or explicit attention-related call was identified.\\
    \botrule
  \end{tabularx}
  \label{tab:8}
\end{table*}

\subsubsection{Error Analysis of the Framework}

In addition to genuine paper-code inconsistencies, the human validation process also identified several cases where the framework produced incorrect predictions. We further summarize these errors into three major categories. For each category, two representative examples are provided in Table~\ref{tab:9}.

\textbf{False Positives Caused by Differences in Code Implement-ation Forms.} The current framework primarily relies on function-level code units for analysis and therefore performs well in identifying explicitly encapsulated implementations. However, in practical software development, not all computational procedures are implemented as independent functions. For example, one project \cite{rosignoli2026deep} described the use of a confusion matrix for metric calculation in the paper. However, the corresponding code was directly embedded within a script rather than implemented as an independent function. Although the functionality was actually present, it was not captured during function-level code extraction, causing the framework to incorrectly classify the sentence as inconsistent. This phenomenon indicates that real-world software often contains diverse implementation patterns, including different levels of abstraction, code reuse, and inline implementations. Therefore, function-level analysis alone may not fully capture implicit implementations. Future approaches should incorporate multi-granularity code representations and program dependency relationships to reduce such false positives.

\textbf{False Negatives Caused by Implicit Methodological Descriptions.} Some paper descriptions involve model training configurations or regularization strategies but are expressed using general terminology or highly abstract descriptions. As a result, the framework may incorrectly match them with semantically similar code implementations. For example, one project \cite{tjaden2026terminatornet} described learning rate settings and L2 regularization strategies, and the framework predicted that these components had corresponding implementations in the code. However, human validation revealed that the corresponding parameter configurations and implementations were not present. This observation suggests that the current training data may still be insufficient to capture diverse forms of methodological descriptions. Future work should incorporate more projects from broader application domains and enrich the diversity of paper-code relationships represented in training data.

\textbf{Insufficient Filtering of Non-Implementation Sentences.} The third category of errors mainly originates from the upstream sentence selection stage. Some paper sentences may involve computational, biological, or AI concepts but essentially represent background introduction, research motivation, or result interpretation rather than specific software functionalities or computational procedures. For example, sentences such as “... to computational and Artificial Intelligence (AI) methods …” or “… metric arises when the model …” do not correspond to specific code implementations. If such sentences are not effectively filtered during preprocessing, the framework may incorrectly treat them as claims requiring code support. This finding demonstrates that paper-code consistency analysis depends not only on cross-modal semantic modeling capabilities but also on the quality of upstream scientific text understanding. Future improvements may incorporate more fine-grained document structure analysis and methodological sentence identification techniques to further enhance analysis accuracy.

\begin{table*}[t]
  \setlength\tabcolsep{7pt}
  \renewcommand{\arraystretch}{1.5}
  \caption{Error types and representative cases observed when applying the consistency framework to real-world bioinformatics software.}
  \centering
  \begin{tabularx}{\textwidth}{@{} >{\hsize=0.7\hsize}X >{\hsize=1.5\hsize}X c >{\hsize=0.8\hsize}X   @{}}
    \toprule
    \multicolumn{1}{c}{\textbf{Type}} & \multicolumn{1}{c}{\textbf{Paper Sentence}} & \textbf{Probability} & \multicolumn{1}{c}{\textbf{Human Validation}} \\
    \midrule
    False Positives Caused by Differences in Code Implementation Forms & From these, we derived accuracy and calculated the area under the ROC curve (AUROC) to assess the model's ability to distinguish between binding and nonbinding proteins.&0.1440&Consistent. The corresponding implementation exists but is not encapsulated as a function. \\
    &Confusion matrices were used to calculate true positives, true negatives, false positives, and false negatives.&0.3195 &Consistent. The corresponding implementation exists but is not encapsulated as a function. \\
    \midrule
    False Negatives Caused by Implicit Methodological Descriptions&To help address possible overfitting, L2 regularization was used with a value of 0.0001.&0.7555& Inconsistent. No L2 regularization setting was identified in the code.\\
    &A constant learning rate was used with a step-size of 0.1.&0.5317&Inconsistent. No learning rate configuration was identified in the code.\\
    \midrule
    Insufficient Filtering of Non-Implementation Sentences&Promising advances have been recently achieved thanks to computational and Artificial Intelligence (AI) methods, possibly opening new avenues for a faster and cheaper option in identifying G4BPs.&0.1137&The statement is descriptive rather than implementation-related.\\
    &However, a limitation of this metric arises when the model incorrectly predicts a large number of unlabeled samples as positive.&0.1618&The statement is descriptive rather than implementation-related.\\
    \botrule
  \end{tabularx}
  \label{tab:9}
\end{table*}

\subsection{Temporal Trend Analysis}

To further investigate how paper-code consistency has evolved in recent years, we extend the temporal scope of our analysis by examining bioinformatics software projects published in \textit{Genome Biology} over the past five years (2022-2026). A total of 89 projects were included in this extended analysis, including 15, 14, 21, 24, and 15 projects published in 2022, 2023, 2024, 2025, and 2026, respectively. Note that the 2026 collection only covers projects published between January and May. Compared with the previous case study, this extended analysis provides continuous yearly coverage and enables a preliminary investigation of temporal variations in paper-code inconsistency.

\begin{figure*}[htbp]
  \centering
  \includegraphics[width=0.98\linewidth]{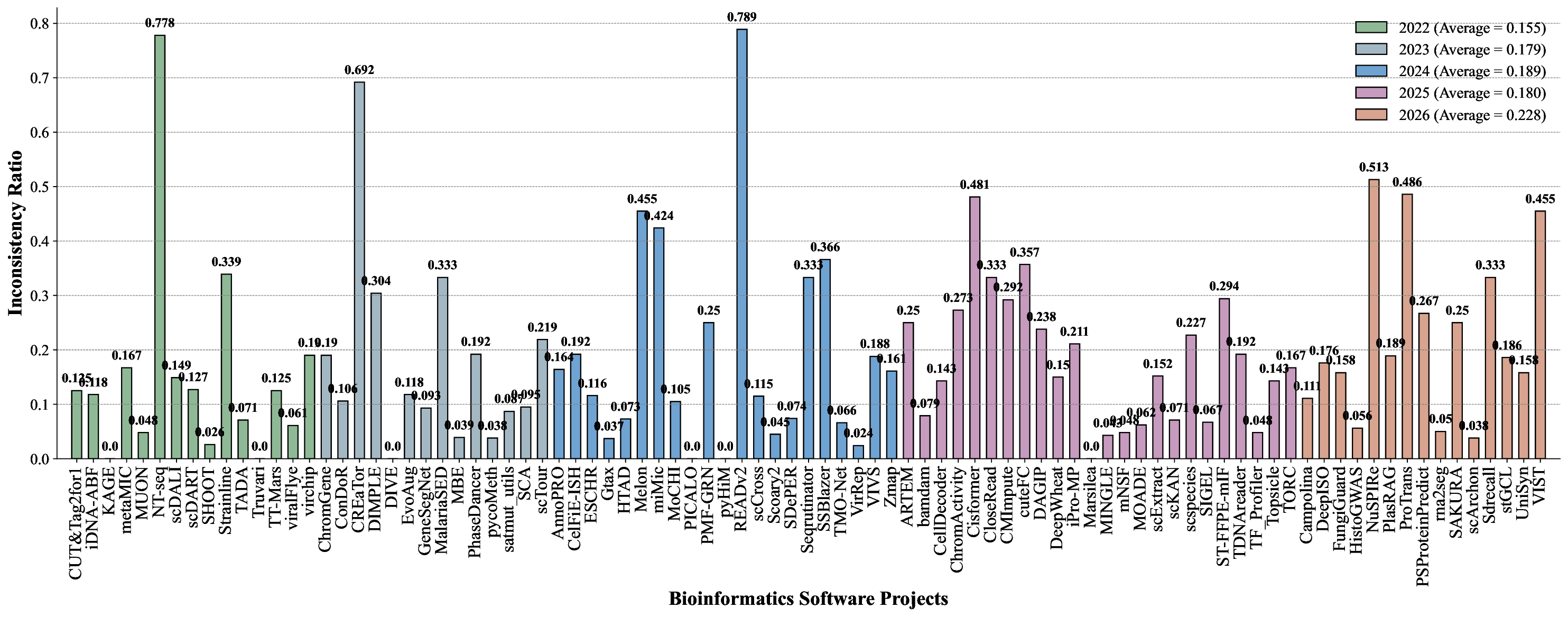}
  \caption{Temporal trends of inconsistency ratios among 89 bioinformatics software projects published in \textit{Genome Biology} from 2022 to 2026. The numbers of projects included in 2022, 2023, 2024, 2025, and 2026 are 15, 14, 21, 24, and 15, respectively. All values are rounded to three decimal places.}
  \label{fig:5}
\end{figure*}

As shown in Figure~\ref{fig:5}, both the number of relevant software projects and the observed inconsistency ratio in Genome Biology exhibit an overall increasing trend over time. Specifically, although fluctuations exist across different years, the average inconsistency ratio increases from 15.5\% in 2022 to 22.8\% in 2026. One possible explanation is the presence of a selection effect in early open-source projects. During periods when public code availability was not yet widely required, researchers who voluntarily released their software may have represented a relatively selective group with greater confidence in software maturity, maintainability, and development quality. In other words, early publicly available projects may not represent the average quality of all software projects developed during that period, but rather projects whose developers were more willing and prepared to share their implementations. This selection effect may have resulted in an underestimation of paper-code inconsistency in earlier years.

With the recent development of open science practices and increasing requirements from journals regarding code availability, the coverage of publicly accessible software has expanded substantially. Consequently, a broader range of projects with different scales, development stages, and levels of maturity have entered the open-source ecosystem. Under this broader inclusion environment, code availability may increase the diversity of analyzable projects and consequently introduce more projects with incomplete documentation, partial mismatches between paper descriptions and implementations, or insufficient code maintenance. This may contribute to the observed increase in inconsistency ratios over time.

However, it is important to emphasize that this temporal trend should not be interpreted simply as evidence that stronger code disclosure requirements have reduced paper-code consistency. Instead, the findings should be understood from a broader perspective: increased code accessibility improves the visibility and verifiability of software implementations, but code accessibility alone does not guarantee consistency between papers and code. Indeed, as more software projects become publicly available, researchers are increasingly able to identify implementation differences, undocumented procedures, and previously hidden gaps that could not be observed when code was inaccessible. Therefore, open-source practices may simultaneously enhance reproducibility and reveal previously invisible paper-code inconsistencies.

Taken together, these findings highlight that as bioinformatics software ecosystems continue to expand, establishing systematic approaches for evaluating the consistency between scientific descriptions and software implementations remains an increasingly important need.

\section{Discussion}
\subsection{Core Findings}

Through a systematic analysis of the consistency relationship between paper descriptions and code implementations, this work reveals an overlooked issue in current bioinformatics software ecosystems: the availability of source code does not necessarily guarantee a clear correspondence between published methodologies, workflows, and experimental procedures and their actual implementations. This issue reflects broader changes in how scientific knowledge is represented and disseminated. In bioinformatics, software is no longer merely a tool for implementing algorithms; instead, it has increasingly become an essential component of scientific contributions. Scientific publications typically serve the roles of methodological explanation and theoretical communication, whereas software code preserves concrete implementation details and computational workflows. Therefore, papers and code together constitute an integrated knowledge representation system in modern bioinformatics. When inconsistencies exist between these two forms of knowledge representation, the interpretability, understandability, and reusability of scientific findings may be affected.

However, the boundary of paper-code consistency is not always absolute. Our case study demonstrates that some training procedures, parameter configurations, and auxiliary analysis scripts may not be publicly released because authors consider them to be non-essential components of the main contribution. Such cases account for 55\% of the observed inconsistencies. Therefore, paper-code consistency should not be simply equated with previous concepts such as “whether code is complete” or “whether experiments are reproducible.” Instead, it involves a broader question regarding how researchers, developers, and users collectively define which types of information constitute essential scientific knowledge. Future bioinformatics communities need further discussion and consensus-building regarding reasonable consistency boundaries for different types of software.

\subsection{Practical Implications for Different Bioinformatics Users}

Paper-code inconsistency not only affects how different user groups understand and utilize bioinformatics software but also influences the broader ecosystem of scientific knowledge dissemination.

For researchers, inconsistencies between paper descriptions and code implementations may increase the cost of reproducing methods and conducting further development. When training procedures or data processing steps cannot be traced through the released code, researchers may need to infer software behaviors independently, potentially leading to incorrect interpretations. This not only reduces the efficiency of reusing existing methods but may also result in subsequent studies being developed based on incomplete information.

For clinical practitioners, bioinformatics software is increasingly being applied in scenarios such as disease prediction, omics analysis, and clinical decision support, where interpretability and reliability are particularly important. If the functionalities described in publications cannot be verified through publicly available implementations, clinical users may find it difficult to assess the applicability, limitations, and potential risks of these tools, thereby affecting their practical adoption.

For students and interdisciplinary researchers, source code has become an important pathway for learning bioinformatics methodologies. Clear correspondence between papers and code can help these users understand algorithmic principles, experimental workflows, and engineering implementations. Conversely, hidden inconsistencies may further increase the learning barriers for newcomers entering the field.

\subsection{Toward FAIR-Oriented paper-code Consistency Guidelines}

In recent years, the FAIR principles (Findable, Accessible, Interoperable, and Reusable) have become important guidelines for promoting scientific data sharing. However, for bioinformatics software, ensuring that code can be discovered and accessed is insufficient to guarantee that it can be effectively understood and reused. We argue that paper-code consistency provides an opportunity to further extend the interpretation of FAIR principles in the context of bioinformatics software.

First, from the perspective of Findability, users should not only be able to locate software repositories but also identify where key methodological descriptions in publications are implemented within the codebase.

Second, regarding Accessibility, publicly available code should not merely be downloadable but should also provide sufficient documentation to help users understand software behaviors and implementation logic.

Third, for Interoperability, clearly defined computational workflows, module relationships, and interface descriptions can facilitate software integration across different research environments.

Finally, in terms of Reusability, complete and traceable implementations provide essential support for method transfer, extension, and secondary research.

Therefore, future efforts toward software openness in bioinformatics should move beyond the question of “whether code is provided” toward the broader goal of “whether scientific implementations are understandable, verifiable, and reusable.” To achieve this goal, journals and research communities could explore lightweight consistency guidelines. For example, authors could provide mappings between methodological descriptions and corresponding code modules; peer-review processes could encourage checking whether critical methodological claims have corresponding implementations; and software repositories could provide clearer version histories and computational workflow documentation. These practices should not be viewed as additional publication burdens, but rather as mechanisms to promote more transparent dissemination of scientific knowledge.

\subsection{Applications and Recommendations}

The framework proposed in this work can not only support large-scale analysis of bioinformatics software ecosystems but also serve as an auxiliary tool in practical software development and evaluation processes.

For researchers, the framework can be incorporated into pre-submission self-check workflows. By providing methodological descriptions from manuscripts, researchers can automatically identify potentially missing or difficult-to-trace implementation components and discover possible issues affecting reproducibility before publication.

For journals and reviewers, the framework can serve as an auxiliary tool for rapidly checking whether key methodological claims in manuscripts have corresponding implementations in publicly available repositories. Importantly, this tool is not intended to replace human review but rather to help reviewers identify potentially problematic areas requiring further examination.

For software maintainers, the proposed approach can also be used to detect long-term divergence between publications and evolving codebases. As software undergoes multiple rounds of updates, published descriptions may gradually become inconsistent with current implementations. Automated consistency analysis can help identify such implementation drift.

For the bioinformatics community as a whole, we believe that one of the most important priorities is to promote continued discussion regarding the boundaries of paper-code consistency. Currently, researchers still hold different interpretations regarding which components constitute essential parts of software and should theoretically be disclosed, and no unified consensus has yet been established. Therefore, we suggest that future communities gradually develop clearer consistency standards through mechanisms such as focused discussions, practical guidelines, and publicly available benchmarks. Compared with pursuing incremental performance improvements on existing tasks, establishing reasonable and actionable consistency standards may ultimately have a more profound impact on the future development of the bioinformatics ecosystem.

\subsection{Limitations}

This work represents the first exploration of paper-code consistency analysis and inevitably has several limitations.

First, BioCon currently only includes bioinformatics software implemented in Python. Although Python has been increasingly adopted in bioinformatics applications, other programming languages, such as R and C/C++, remain widely used in the field. Therefore, the current findings may not fully represent the characteristics of all types of bioinformatics software. Future work should extend BioCon to support multiple programming languages and enable more comprehensive investigations.

Second, the current analysis operates at the granularity of paper sentences and function-level code implementations. However, bioinformatics software often contains complex architectures in which a complete functionality may be achieved through interactions among multiple modules, or implementations may exist without being explicitly encapsulated as independent functions. Consequently, function-level analysis may still introduce detection errors. Future work will incorporate additional code structures, function dependency relationships, and documentation information to improve the analytical capability and accuracy of the framework.

Finally, although the proposed framework substantially reduces the cost of large-scale analysis, the current results still require interpretation with human validation. Future research will explore more advanced cross-modal consistency analysis approaches to improve accuracy, robustness, and interpretability when applied to increasingly complex software systems.

\section{Conflicts of interest}
The authors declare that they have no competing interests.

\section{Funding}
This work was supported by the National Natural Science Foundation of 
China under Grant Nos. 72274152, 62402376, and 62572389.

\section{Data availability}
The complete source code for BioCon construction, model training, validation, and testing is publicly available at \url{https://github.com/Ricardo1998-Xu/BioCon}. The complete BioCon dataset is available at \url{https://github.com/Ricardo1998-Xu/BioCon/blob/main/data/BioCon/BioCon.jsonl}. The datasets and analysis results for the case study, including the corresponding publications, code repositories, and consistency analysis results, are available at \url{https://github.com/Ricardo1998-Xu/BioCon/tree/main/Case%20Study}.

\section{Author contributions statement}
T.X. and J.W. conceived the experiments. T.X., X.L., S.D. and H.C. curated the data. T.X. and X.Z. conducted the experiments. T.X. and X.L. analysed the results. T.X., X.Z. and J.W. wrote and reviewed the manuscript.

\bibliographystyle{unsrt}
\bibliography{reference}



\section{Figure Legends}
Figure 1 caption: Overview of the BioCon benchmark construction pipeline.

Alt text: Flowchart illustrating the main steps for constructing the BioCon benchmark, from bioinformatics project and paper collection to sentence-code pair generation and sample generation.

Figure 2 caption: Overall architecture of the proposed cross-modal consistency analysis framework.

Alt text: Schematic diagram showing the framework for analyzing consistency between scientific paper statements and corresponding code implementation units.

Figure 3 caption: Cumulative distribution function curves of ranking results. Panels (a) and (b) present the CDF curves of five models under the intra-project and cross-project settings, respectively. In each CDF curve, the x-axis represents the ranking position, while the y-axis indicates the cumulative proportion of queries whose retrieved implementation units have ranks less than or equal to the corresponding value.

Alt text: Two panels show cumulative distribution function curves for five models, comparing ranking performance under intra-project and cross-project settings. In each CDF curve, the x-axis represents the ranking position, while the y-axis indicates the cumulative proportion of queries whose retrieved implementation units have ranks less than or equal to the corresponding value.

Figure 4 caption: Inconsistency ratios across 44 real-world bioinformatics software projects. The selected projects were collected from the latest issues of Bioinformatics, Genome Biology, and PLOS Computational Biology, including 23, 15, and 6 projects, respectively. All values are rounded to three decimal places.

Alt text: Plot showing the inconsistency ratios of 44 real-world bioinformatics software projects, with each project represented by its corresponding inconsistency ratio.

Figure 5 caption: Temporal trends of inconsistency ratios among 89 bioinformatics software projects published in \textit{Genome Biology} from 2022 to 2026. The numbers of projects included in 2022, 2023, 2024, 2025, and 2026 are 15, 14, 21, 24, and 15, respectively. All values are rounded to three decimal places.

Alt text: Plot showing the yearly distribution of inconsistency ratios among bioinformatics software projects published in \textit{Genome Biology} from 2022 to 2026, illustrating changes in inconsistency over time.

\section{Biographical Note}
Tianxiang Xu is currently working toward the Ph.D. degree with the School of Computer Science and Technology, Xi’an Jiaotong University, China. His research interests include biomedical software engineering, focused on defect prediction and localization.

Xiaoyan Zhu received her Ph.D. degree in computer science and technology from Xi’an Jiaotong University, China, in 2013. Currently, she is working as an associate professor in the School of Computer Science and Technology, Xi’an Jiaotong University. Her research interests include software engineering, data mining and machine learning.

Xin Lai received the BS degree and the PhD degree from University of Science and Technology of China in 2003 and 2009, respectively. Meanwhile, he also received a joint-PhD degree from City University of Hong Kong in 2009. He is currently an Associate Professor with Xi’an Jiaotong University. His research interests include medical risks, long-term survivor models and risk-adjusted monitoring approaches.

Xin Lian is currently studying for a Ph.D. degree in the School of Computer Science and Technology in Xi’an Jiaotong University. Her research interests include bioinformatics, statistical process control and software testing.

Sizhe Dang is currently pursuing the Doctor of Engineering degree in Computer Science and Technology at Xi’an Jiaotong University. His research interests include machine learning, optimization, computer vision, and multimodal learning.

Hangyu Cheng is currently working toward the Ph.D. degree with the School of Computer Science and Technology, Xi'an Jiaotong University, China. His research interests include biomedical software engineering, focused on program analysis and repair.

Jiayin Wang received the B.S. degree in computer science and technology from Xi’an Jiaotong University, Xi’an, China, in 2008 and the Ph.D. degree in computer science and engineering from University of Connecticut, Storrs, USA, in 2013. From 2013 to 2015, he was a Postdoctoral Research Associate with The Genome Institute at School of Medicine, Washington University in St. Louis, USA. Since 2015, he has been a Professor with Department of Computer Science and Technology, Xi’an Jiaotong University. His research interest includes the algorithms and systems for the applications on bioinformatics, computational biology and cancer genomics.

\end{document}